\documentclass[letterpaper, 10 pt, conference]{ieeeconf}  

\IEEEoverridecommandlockouts                              

\usepackage{graphicx}
\usepackage{amsmath}
\usepackage{amssymb}
\usepackage{booktabs}
\usepackage{marvosym}
\usepackage{multirow}  
\usepackage{makecell}  
\usepackage{subcaption}
\usepackage{colortbl}

\title{\LARGE \bf
Boosting Online 3D Multi-Object Tracking through Camera-Radar Cross Check
}

\author{Sheng-Yao Kuan$^{1}$, Jen-Hao Cheng$^{2}$, Hsiang-Wei Huang$^{2}$, Wenhao Chai$^{2}$, Cheng-Yen Yang$^{2}$, \\Hugo Latapie$^{3}$, Gaowen Liu$^{3}$, Bing-Fei Wu$^{1}$, Jenq-Neng Hwang$^{2}$
\thanks{$^{1}$Sheng-Yao Kuan and Bing-Fei Wu are with National Yang Ming Chiao Tung University, Taiwan. \tt\small\{shaunkuan.10,bwu \}@nycu.edu.tw
        }
\thanks{$^{2}$Jen-Hao Cheng, Hsiang-Wei Huang, Wenhao Chai, Cheng-Yen Yang and Jenq-Neng Hwang are with University of Washington, USA. \tt\small \{andyhci, , hwang\}@uw.edu
        }
\thanks{$^{3}$Hugo Latapie and Gaowen Liu are with Cisco, USA. \tt\small \{hlatapie, gaoliu\}@cisco.com
        }
}

\begin{document}

\maketitle
\thispagestyle{empty}
\pagestyle{empty}


\begin{abstract}
In the domain of autonomous driving, the integration of multi-modal perception techniques based on data from diverse sensors has demonstrated substantial progress. Effectively surpassing the capabilities of state-of-the-art single-modality detectors through sensor fusion remains an active challenge. This work leverages the respective advantages of cameras in perspective view and radars in Bird's Eye View (BEV) to greatly enhance overall detection and tracking performance.
Our approach, Camera-Radar Associated Fusion Tracking Booster (CRAFTBooster) represents a pioneering effort to enhance radar-camera fusion in the tracking stage, contributing to improved 3D MOT accuracy. The superior experimental results on K-Radaar dataset, which exhibit 5-6\% on IDF1 tracking performance gain, validate the potential of effective sensor fusion in advancing autonomous driving. 

\end{abstract}

\section{Introduction}

    

\begin{figure*}[t]
    \centering
    \includegraphics[width=1\linewidth]{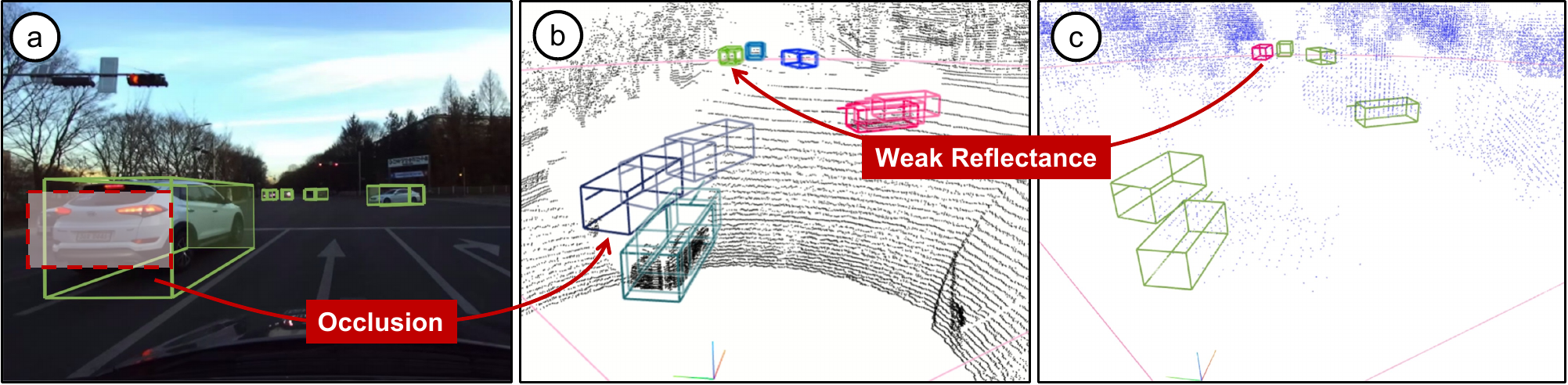}
    \caption{The image-based detections frequently confront challenge related to occlusion as shown in (a), whereas the radar-based detections encounter complications due to weak reflectance as shown in (c). The tracking result shows in (b) is from our method, \textbf{CRAFTBooster}, which can  address these issue by fusion in tracking stage. Note that the background in (b) is LiDAR point cloud and the one in (c) is radar point cloud generated by the Constant False Alarm Rate (CFAR) process and both of these data are only for demostration purpose.}
    \label{fig:cover}
\end{figure*}


Similar to the role of eyes in the field of autonomous driving, perception serves as the cornerstone for subsequent functions like motion prediction, path planning, and maneuver control. 
Multi-modal perception techniques that integrate data from different sensors (such as camera, LiDAR, radar, etc) have shown significant progresses in recent years. Recently, most of these researches leverage the robust representational capabilities of Neural Networks (NNs) to extract intricate semantic features from the input \cite{liu2022bevfusion,liang2022bevfusion}. However, features generated by multi-sensor neural networks can be susceptible to sensor noise, stemming from possible sensor defects and inadequate calibration \cite{Zhu_2023_CVPR}. Therefore, identifying an effective multi-sensor fusion approach that surpasses the capabilities of state-of-the-art single-modality detectors is challenging.

    

Tracking of dynamic objects around the vehicle is crucial to autonomous navigation, including path planning and obstacle avoidance. Multiple Object Tracking (MOT)~\cite{huang2024exploring} requires the precise identification and localization of dynamic objects in the surrounding environment of the ego car. Obstructions and interactions among objects with similar appearances are two primary factors that render MOT a challenging task. Due to the wealth of semantic information, cameras are extensively employed for tasks such as object detection and tracking. While radars, known for achieving higher spatial resolution and interference reduction, are commonly employed in Advanced Driver Assistance Systems (ADAS) like collision avoidance~\cite{CNN_RUDet_3DRadarCube}, showcasing highly robust and reliable in adverse weather conditions such as fog and snow. Our approach leverages the advantages of the camera in the perspective view and the radar's ability to highlight depth accuracy in the bird's eye view (BEV) to enhance the tracking performance.

Compared with previous sensor fusion endeavors, this work represents a novel effort to enhance the efficacy of radar-camera fusion in the tracking stage, consistently improving 3D MOT accuracy. To be specific, 
\begin{itemize}
    \item We propose a novel fusion strategy at tracking level, named Camera-Radar Associated Fusion Tracking Booster (CRAFTBooster), which further enhances performance by cross-check between the tracking results of both sensors.
    \item Considering the attributes of camera and radar, we simultaneously evaluate representations in both perspective view and BEV. This entails confirmation of unreliable detection results from each modality.
    \item Extensive experiments are conducted based on multiple trackers with same detectors, and gain around
5-6\% IDF1 improvements on Kradar and 1-2\% on CRUW3D benchmark.
\end{itemize}

\begin{figure*}[t]
    \centering
    \includegraphics[width=1\linewidth]{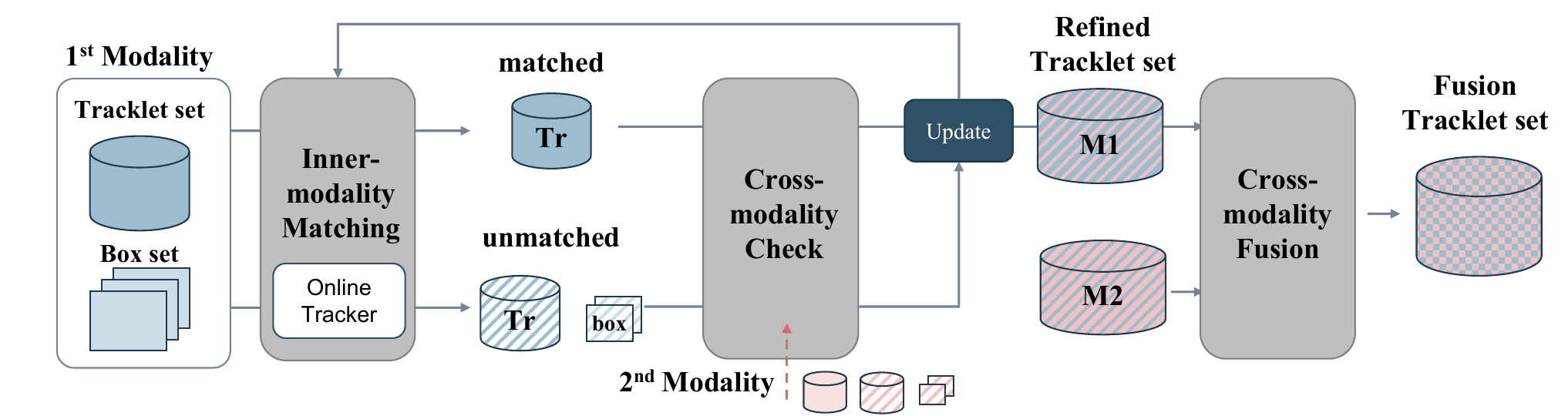}
    \caption{CRAFTBooster is a comprehensive multi-modality fusion framework designed for the 3D MOT task based on camera and radar, specifically utilizing detection results from 3D object detections in two modalities as its input. The architecture is comprised of three main components: Inner-modality Matching Module, Cross-modality Check Module and Multi-modality Fusion Module. Tr. denotes tracklet and Det. denotes detections. Modality 1\(^{st}\) and 2\(^{nd}\) can alternate between camera and radar.}
    \label{fig:pipline-CrossCheck}
\end{figure*}

\section{Related Works}
\subsection{Multi-Modality 3D Perception}
Accurate perception is crucial for prediction and planning tasks in full-stack autonomous driving systems \cite{uniad}. 
Three primary sensing modalities, LiDAR, radar, and camera, exhibit unique strengths and weaknesses. 
LiDAR is renowned for its geometric reasoning capabilities \cite{pvrcnn, centerpoint}, outperforming camera and radar in 3D object detection. On the other hand, camera-based methods \cite{yuhong-CMKD-ECCV2022}, though lack of reliable depth information, have excelled in providing rich semantic and dense pixel information, pivotal for scene understanding, at a lower cost, facilitating broader deployment. Unfortunately, cameras also suffer from adverse weather or lighting conditions.
Robust all-weather perception is essential for the reliability of autonomous driving systems \cite{paek2023kradar}. 
Millimeter-wave radar, though lacks in semantic information of detecting objects, with its resilience to poor lighting and adverse weather conditions \cite{9607427, cheng2023centerradarnet}, emerges as a valuable complement to LiDAR and camera. 
It offers a promising pathway to a robust, cost-effective 3D perception system, where radar serves as a supplementary modality along with cameras.

Multi-modality fusion capitalizes on the complementary strengths of different sensors to achieve more accurate and reliable predictions~\cite{chai2022deep}. 
Depending on the adoption of fusion in the algorithmic pipeline stages, methods are categorized as early \cite{Vora2019PointPaintingSF, Huang2020EPNetEP}, deep \cite{TransFusion, liang2022bevfusion, bevfusion}, or late fusion \cite{CLOCs}. 
These techniques have been developed and assessed in ideal conditions with high-quality data.
However, recent studies \cite{fusion_rob_benchmarking, 3d_rob_benchmarking} reveal that existing fusion approaches are susceptible to common real-world corruptions, even with data augmentation during training. 
Such corruptions often occur when a modality is compromised, for instance, due to harsh weather or sensor malfunctions. 
Incorporating radar as one of the fusion modalities has been proposed as a solution to enhance robustness.
Recent research has focused on improving fusion robustness by merging features from images and radar point clouds \cite{nabati2021centerfusion, simplebev}. 
Yet, these methods face two primary challenges. 
First, the information derived from radar points is less than that from radar tensors \cite{rodnet}, limiting detection capabilities. 
Second, these methods rely heavily on a single modality, making them vulnerable to severe corruption in the primary modality \cite{3d_rob_benchmarking}.
To address these challenges, we introduce a new fusion approach that utilizes dual streams of 3D detection and tracking information from camera images and radar tensors. 
This method refines each modality online while preserving its independence. 
Unlike previous approaches, our method operates without training and demonstrates its robustness and adaptability in scenarios challenging for cameras or radar.

\subsection{Online Tracklet Booster Methods}
Tracklet booster modules for enhancing tracking performance are divided into online and offline methods. 
Online methods offer immediate improvements by integrating existing tracking approaches, while offline methods use global information for more substantial enhancements.\\
Various online tracklet booster methods add an extra association stage to recover erroneous (false positive or false negative) detections. For instance, Observation Centric-SORT \cite{cao2023observationcentric} introduces a second, position-based association stage to improve tracking. Peng et al. \cite{du2022giaotracker} propose box-plane matching and use historic-tracklet appearance data for robust association. CAMO-MOT creates an optimal occlusion State-based Object Appearance Module~(O2S-OAM) to use occlusion head to identify occlusion state on each bbx, which emphasizes the significance of occlusion recognition. Moreover, some approaches, like \cite{cai2022memot,song2023moviechat}, employ a memory buffer for long-term tracklet re-identification, reducing ID switches due to occlusion or camera re-entry.\\

Despite their effectiveness, they predominantly rely on single-modality information. To address this limitation, we introduce a novel online tracklet cross-check module. This plug-and-play solution significantly enhances tracking performance by leveraging multi-modality attributes and integrating them seamlessly into the tracking process.

\section{Proposed Method}  \label{section_method}

\subsection{Overview}
The motivation behind the CRAFTBooster is straightforward but highly effective. By maintaining the independence of each sensor's operation, we aim to extract valuable information from both modalities to significantly improve overall performance. Fig~\ref{fig:pipline-CrossCheck} presents the main framework of this work. There are three main modules in this pipeline. 

\subsection{Inner-modality Matching Module}
By employing high-quality detectors, we can obtain reliable detection results from the raw data of each sensor. Then we choose a proficient online tracker to perform the 3D MOT task on both modalities.
In this work, we utilize BOTSORT~\cite{aharon2022botsort} or Bytetrack \cite{zhang2022bytetrack} as the online tracker and make necessary modifications, such as removing the Camera Motion Compensation~(CMC) module. Note that this work is not confined to a specific online tracker, one can utilize other alternatives to extract good trajectories.

Our task revolves the ambiguities created by 3D MOT results from both modalities. Therefore, the definition of detection is as \(x=[x,y,z,l,w,h,sin\theta,cos\theta]\), where \((x_c,y_c,z_c)\) is the 3D coordinate of the bounding box center, \((l,w,h)\) is the size of bounding box. \(\theta\) is the orientation along with axis z, so \((sin\theta,cos\theta)\) are used to describe the heading rotation. We replace \(\theta\)  with \(sin\theta\) and \(cos\theta\), because the prediction would encounter issues when the value of \(\theta\) approaches 0.

One essential criterion is that this tracker must be capable of extracting matched tracklets, unmatched tracklets and the rest detection bounding boxes that couldn't be matched.

\subsection{Cross-modality Check Module}
This module constitutes a critical component of the entire work. The main task include two primary objectives. Firstly, recovering lost information in those unmatched tracklets, which do not get assigned with the detections in current frame, by leveraging pertinent data from the other modality. Secondly, conducting an evaluation of unmatched detections, which cannot be assigned to existing tracklets, from a different perspective through the other modality to ensure the practicality. The details of these tasks will be presented in following sections. This module outputs the refined tracklet set for each modality.

\paragraph{Cross-modality Check on Unmatched Tracklets.}

\begin{figure}[t]
    \centering
    \includegraphics[width=1\linewidth]{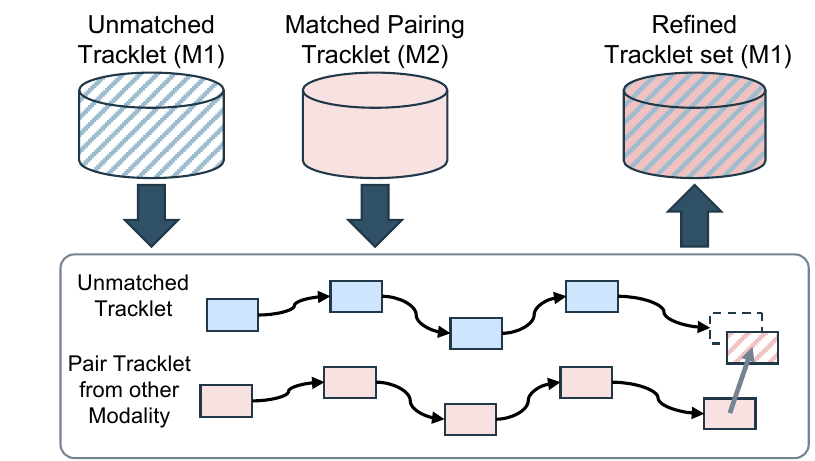}
    \caption{\textbf{Cross-modality Check on Unmatched Tracklets}. The missing data in unmatched tracklets would be recovered by the paired and active tracklet information.}
    \label{fig:CCM_Tracklet}
\end{figure}

Sensor limitations and environmental factors, like different lighting or weather conditions, can cause abrupt data loss during object tracking on roads. Although some trackers can address this with motion models, extended periods of data loss may result in inaccuracies and ID switch issues. 

The concept of this Cross-modality Check Module is shown in Fig \ref{fig:CCM_Tracklet}. The inputs are unmatched tracklets, which are from the Inner-modality Matching module and also the paired tracklets, which are from the other modality. The pairing method will be discussed in \textbf{Multi-modality Fusion module}. We utilize the detections from the paired and active tracklet to fill in missing data. Subsequently, the motion model of the online tracker will be updated later.

\paragraph{Cross-modality Check on Unmatched Detections.}

\begin{figure}[t]
    \centering
    \includegraphics[width=1\linewidth]{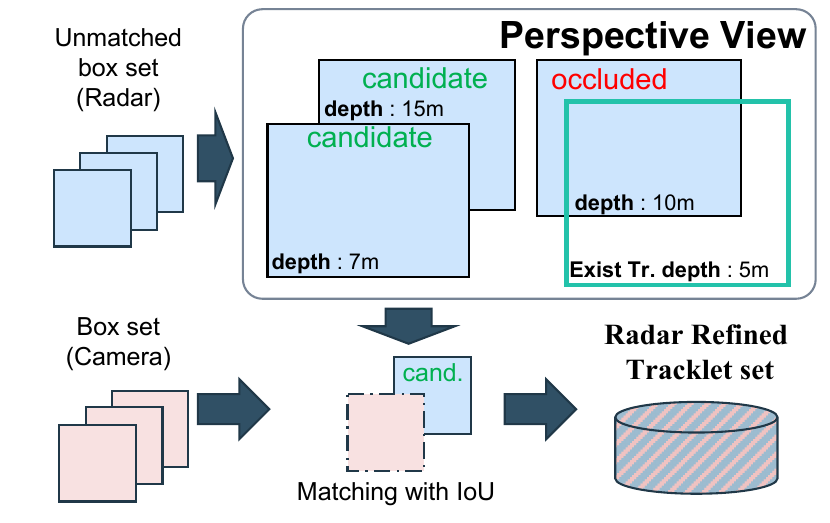}
    \caption{\textbf{Cross-modality Check on Unmatched Detections} in Perspective View. The unmatched detections from radar would be projected to perspective view to be checked with camera detections.}
    \label{fig:CCM_Dets_FrontView}
\end{figure}

\begin{figure}[t]
    \centering
    \includegraphics[width=1\linewidth]{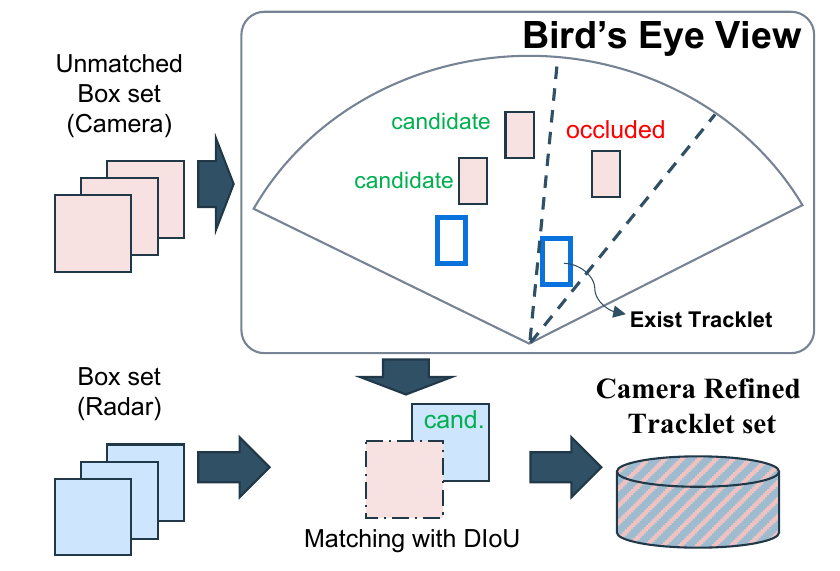}
    \caption{\textbf{Cross-modality Check on Unmatched Detections} in BEV. The unmatched detections from camera would be projected to BEV to be checked with radar detections.}
    \label{fig:CCM_Dets_BEV}
\end{figure}

In some cases, some detections may not be successfully matched with established tracklets. 
To address this, we perform another check to decide whether to keep these objects. 
Since different modalities have detectors optimized for their respective advantageous spaces, we assess these scenarios from different perspectives for different sensors. 
Cameras give detailed frontal views but struggle with occlusive objects, while radar sensors excel in depth accuracy and offer better results from BEV.
To address this, we consider objects' occlusion states in the camera's perspective view and radar's BEV.
First, we project the established tracklets and the unmatched bounding boxes from radar detections onto the perspective view. 
Next, we assess whether there is occlusion between the established tracklets and the unmatched bounding boxes. If they overlap, we discard the unmatched bounding boxes.
Due to uncertainty, unmatched bounding boxes do not necessarily occlude each other. 
Thus, they are counted as candidates for the next matching stage with camera's detections. 
As shown in Fig~\ref{fig:CCM_Dets_FrontView}, we employ IoU as the matching criterion for radar candidates and camera's detections. 
A successful match creates a new tracklet, updating the radar tracker's motion model state.
\begin{equation}\label{eq:DIOU}
DIoU= 1 - IoU + \frac{\rho^2(b_1,b_2)}{c^2}.
\end{equation}
Likewise, we project unmatched camera's detections onto BEV as shown in Fig~\ref{fig:CCM_Dets_BEV} and follow a similar principle in Fig~\ref{fig:CCM_Dets_FrontView} to update the camera's online tracker. 
Unlike the previous matching stage, we opt for a distance-based similarity measure named \(DIoU\)~\cite{zheng2019distanceiou}.
Due to the camera's lower accuracy in depth estimation, IoU encounters challenges in effective matching. 
On the contrary, \(DIoU\) minimizes the normalized distance between central points of two bounding boxes, \(b_1\) and \(b_2\) in Eq~\ref{eq:DIOU}. \(\rho(\cdot)\) is the Euclidean distance, and \(c\) is the diagonal length of the smallest enclosing box covering the two boxes. 

\subsection{Multi-modality Fusion Module}
This module consists of two steps including pairing tracklets and fusing the successfully paired tracklets to generate new tracklets. In actual driving scenarios, we cannot determine which sensor's results are more accurate through model predictions. 
Therefore, this module attempts to further fuse the tracklets refined by Cross-modality Check module from two modalities rationally and efficiently. 

\begin{figure}[t]
    \centering
    \includegraphics[width=1\linewidth]{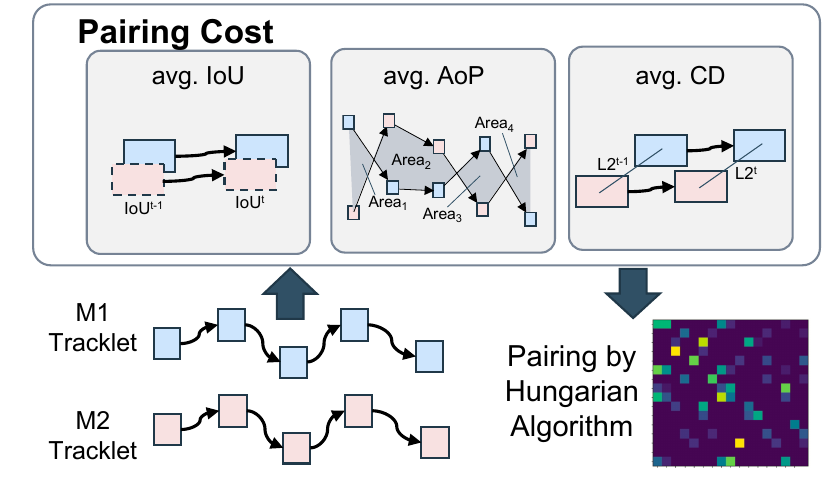}
    \caption{\textbf{Tracklet Pairing} with different methods, avg IoU, avg AoP and avg CD.}
    \label{fig:MultiMFusion_Pair}
\end{figure}

\paragraph{Tracklet Pairing.}
The choice of pairing method significantly affects the Cross-modality Check module's ability to fill missing data. 
We compare three pairing methods: average IoU, average Area of Polylines (AoP), and average Center Distance (CD), as depicted in Fig~\ref{fig:MultiMFusion_Pair}. 

Each tracklet retains location information from the previous \(n\) frames (set to 5 in our experiments). The pairing cost can be determined by calculating the average of IoU among the last \(n\) frames for all tracklets. Next, the average of AoP method considers the areas enclosing the points of two tracklets; a larger area indicates a greater distance between them. Alternatively, the average of CD evaluates the \(L2\) distance between locations. 
The pairing tracklet set is denoted as \(Tr^f_i=\{(Tr_{i1},Tr_{i2}),...\}\). \(Tr_{i1}\) and \(Tr_{i2}\) represent the tracklets from modality \(i1\) and \(i2\) (camera and radar), respectively, within the paired tracklet \(i\).

\paragraph{Tracklet Fusion with Homoscedastic Uncertainty.}
 This module tries to generate a fused tracklet set, which performance can be close to the better refined tracklet set among two modalities. Because we cannot guarantee which modality performs better without additional information. Our goal is to assess the uncertainty of each paired tracklet \(i\) and treat it as the fusion weight of the new location \(X^f_i\).
Firstly, we assume a relatively consistent variability in location displacement for objects on the road within consecutive frames. Consequently, the standard deviation of it is chosen as an indicator to evaluate uncertainty. 
\begin{figure}[t]
    \centering
    \includegraphics[width=0.9\linewidth]{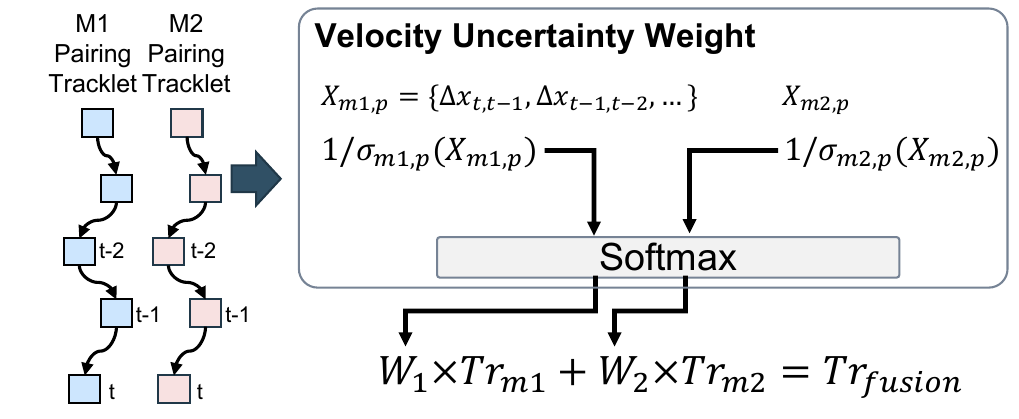}
    \caption{\textbf{Tracklets Fusion with Homoscedastic Uncertainty}. 
 The paired tracklet is fused by velocity uncertainty weighting.}
    \label{fig:MultiMFusion_Fuse}
\end{figure}

The new location for each paired tracklet \(i\) is determined in Eq~\ref{eq:paired_fusion}. \(X_{i1,t}\) \(X_{i2,t}\) represents the location displacement set of modality \(i1\) , \(i2\) at the frame \(t\) respectively. For example, \(X_{i1,t}=\{\Delta x_{i1,t},\Delta x_{i1,t-1},...,\Delta x_{i1,t-n}\}\). \(\Delta x_{i1,t}\) means the distance between the location \(x_{i1,t}\) and \(x_{i1,t-1}\).

\begin{equation}\label{eq:paired_fusion}
X^f_{i,t}=W_{i1,t} \times X_{i1,t}+W_{i2,t}\times X_{i2,t},
\end{equation}
, where \(W_{i1,t},W_{i2,t}\) are the weights defined in Eq~\ref{eq:paired_weight}. 
We employ the \(softmax\) function to align fusion results more closely to the more reliable modality. In Eq~\ref{eq:paired_std}, we calculate the standard deviation of location displacement as the score.

\begin{equation}\label{eq:paired_weight}
W_{m,t} =1/{(\frac{e^{\sigma_{m,t}}}{e^{\sigma_{i1,t}}+e^{\sigma_{i2,t}}}) }\quad \forall m \in \{i_1,i_2\},
\end{equation}
, where 
\begin{equation}\label{eq:paired_std}
\sigma_{m,t} = \frac{1}{n-1} \sqrt{ \sum_{j=t-n}^t (\Delta x_{m,j} - \bar{x_m})^2} \quad \forall m \in \{i_1,i_2\}.
\end{equation}

\section{Experiments} \label{section_exp}

\paragraph{Datasets and Evaluation Metrics.}

We evaluate our method on the K-Radar~\cite{paek2023kradar} and CRUW3D~\cite{cruw3d} datasets as they uniquely provide radar tensors and camera video frames with 3D object bounding box and tracking annotations in diverse driving scenarios. 
We do not use other databases solely containing radar point clouds, such as nuScenes\cite{nuscenes}, TJ4DRadSet\cite{Zheng_2022} and RadarScenes\cite{schumann2021radarscenes}. 
As indicated by the results on these datasets, the performance of 3D perception using radar point clouds is not comparable with that of using the camera unless high-end radar equipment like RADIATE\cite{sheeny2021radiate} is employed. 
K-Radar encompasses scenarios, characterized by varying illumination, weather conditions, and road structures. 
It offers synchronized data, including LiDAR point clouds, 4D radar tensors, and camera images. 
It is divided into approximately 17.5k training and 17.5k testing frames from its raw continuous sequences.
CRUW3D comprises 66K synchronized camera frames, 3D radar tensors, and LiDAR point clouds in different lighting conditions. 
In CRUW3D, we select ten scenarios which contain 18K frames in total, and divide half of the frames for training and the other half for testing.
Table~\ref{tab:trackerPerformance} shows that there is a marginal difference in camera's performance between the two datasets. 
We attribute this observation to the different characteristics of the two datasets. 
Compared to K-Radar, CRUW3D's driving scenes are in good weather, which favors camera-based perception methods.
We adopt MOTA and IDF1 in evaluation to measure 3D object tracking performance.

\paragraph{Detectors.}
 Following the paradigm of tracking by detection, 
 we implement advanced monocular and radar-based 3D object detectors to lay a firm foundation for the subsequent tracking stage. 
 We choose DD3D~\cite{park2021pseudolidar} as our camera's baseline detector, as it is renowned for its superiority in monocular 3D object detection owing to its strong depth-inferring ability. 
 DD3D provides two backbones pre-trained on large-scale depth images~\cite{packnet}.
 We use the DLA-34~\cite{8578353} pre-trained backbone and fine-tune the detector on the K-Radar and CRUW3D datasets. 
 Since DD3D does not infer re-ID embedding features, we apply the publicly available YOLOX\cite{ge2021yolox} detector to extract re-ID features from cropped image regions projected by DD3D's detected 3D bounding boxes. 
 We pick CenterRadarNet~\cite{cheng2023centerradarnet} as our radar's baseline 3D object detector. 
 CenterRadarNet is a joint architecture for 3D object detection and tracking. 
 It learns discriminative representation from 4D radar tensors and favors object localization and re-identification (re-ID) tasks. 
 We employ the same detectors across all experiments. 

\paragraph{Results.} Table~\ref{tab:trackerPerformance} presents the performance of methods on K-Radar and CRUW3D. We use several MOT algorithms for comparison. In CRAFTBooster, we employ ByteTrack and BOTSORT as online trackers within the \textbf{Inner-modality Matching Module}. 
The results demonstrate that our model enhances the performance of the camera's and radar's baseline models. 
It is worth mentioning that the K-Radar encompasses numerous adverse weather conditions, where vehicle objects are entirely obscured in the images. 
Thus, the performance of the camera's perception model, especially compared with the result from the radar's perception model, is significantly lower. 
In our experiments, CRAFTBooster performs best when utilizing BOTSORT as the online tracker in \textbf{Inner-modality Matching Module}. 
Note that our approach surpasses the better modality's performance by 3.6\% to 4.8\% in IDF1.

\begin{table}[]
    \centering
    \resizebox{1\linewidth}{!}{
    \begin{tabular}{cc cc|cc}
        \toprule
        \multirowcell{2}{Tracking\\Method}&\multirowcell{2}{Modality}&\multicolumn{2}{c}{K-Radar}&\multicolumn{2}{c}{CRUW3D}\\
        & & IDF1$\uparrow$ & MOTA$\uparrow$ & IDF1$\uparrow$ & MOTA$\uparrow$ \\
        \midrule

        \multirow{2}{*}{OC-SORT\cite{cao2023observationcentric}}& Camera & 39.0 & 8.4 & 70.6 & 69.2 \\
        &Radar & 56.0 & 36.0 & 51.8 & 41.3 \\
        \midrule
        
        \multirow{2}{*}{StrongSORT*\cite{du2023strongsort}}& Camera & 40.3 & 18.2 & 69.8 & 69.2 \\
        &Radar & 51.5 & 38.1 & 51.9& 41.6 \\
        \midrule
        
        \multirow{2}{*}{ByteTrack\cite{zhang2022bytetrack}}& Camera & 41.4 & 19.1 & 71.3 & 69.7  \\
        &Radar & 52.5 & 42.5 & 53.5 & 41.8 \\
        \rowcolor[gray]{0.9}
        CRAFTBooster & Radar+Camera & 58.3 &44.7& 72.0 & 69.8 \\
        \midrule
        
        \multirow{2}{*}{BOTSORT\cite{aharon2022botsort}}&Camera & 45.3& 20.0 & 71.1 & 69.4 \\
        & Radar & 55.3 & 41.1 & 50.6 & 39.6 \\
        \rowcolor[gray]{0.9}
        \textbf{CRAFTBooster}& Radar+Camera & \textbf{59.5}& \textbf{44.9} & 74.2 & 69.4 \\

        \bottomrule
    \end{tabular}}
    \caption{Overall Performance on K-Radar and CRUW3D Dataset. We implement our method on both ByteTrack and BOTSORT. StrongSORT*: We implement StrongSORT with some modifications, such as removing AFLink, ECC and also BoT( We replace it with our own detector).}
    \label{tab:trackerPerformance}
\end{table}

Our method enables each modality to obtain a refined tracklet set after \textbf{Cross-modality Check Module}. 
This tracklet set can be viewed as the outcome of enhanced performance complemented by the other modality. 
In Table~\ref{tab:CR_Refined}, we discuss the results of camera and radar separately. 
In both the radar and camera refined tracklets, we observe an increase of 3-8\% in IDF1 compared to performance of the original tracker. From the number of false negatives (FNs), it is evident that our module effectively recovers the missing data, although it may inveitably introduce some invalid data in false positives (FPs). It is also worth mentioning that the performance of the refined tracklet set can even surpass the final results of CRAFTBooster. The primary goal of the \textbf{Multi-modality Fusion Module} is to identify a comparable modality that aligns closely with the better-performing sensor modality in scenarios where additional information about sensor conditions is lacking. 

\begin{table}[!h]
    \centering
    \resizebox{1\linewidth}{!}{
    \begin{tabular}{ccccccc}
        \toprule
         \makecell[c]{Modality}&\makecell[c]{Tracking\\Method}& \makecell[c]{IDF1$\uparrow$} & \makecell[c]{MOTA$\uparrow$\\ }&FP$\downarrow$&FN$\downarrow$&IDs$\downarrow$\\
        \midrule
        \multirowcell{6}{Radar}&OC-SORT\cite{cao2023observationcentric}& 56.0 & 36.0 & 4,554 & 14,327 &640\\ 
        &StrongSORT*\cite{du2023strongsort} & 51.5& 38.1 & 3,675 & 15,054 &422\\ 
        &ByteTrack\cite{zhang2022bytetrack} & 52.5 & 42.5 & 2,557 & 14,824 & 415 \\
        &BOTSORT\cite{aharon2022botsort}& 55.3 & 41.1 & 2,180 & 15,640 & 423 \\
        \rowcolor[gray]{0.9}
        &\textbf{Ours w/ ByteTrack}& 61.1 & 46.9 & 3,212 & 12,941 & 305 \\
        \rowcolor[gray]{0.9}
        &\textbf{Ours w/ BOTSORT}& \textbf{63.4} & \textbf{47.2} & 3,316 & 12,775 & 266 \\
        \midrule
        \multirowcell{6}{Camera}&OC-SORT\cite{cao2023observationcentric}& 39.1 & 8.4 & 8,725 & 18,332 & 1,332\\ 
        &StrongSORT*\cite{du2023strongsort} & 40.3 & 18.2& 3,541 & 21,341 &425\\ 
        &ByteTrack\cite{zhang2022bytetrack} & 41.4 & 19.1 & 6,335 & 18,338 & 389 \\ 
        &BOTSORT\cite{aharon2022botsort}& 45.3 & 20.0 & 5,824 & 18,648 & 306 \\
        \rowcolor[gray]{0.9}
        &\textbf{Ours w/ ByteTrack}& 46.9 & 21.2 & 6502 & 17,579 & 315 \\
        \rowcolor[gray]{0.9}
        &\textbf{Ours w/ BOTSORT}& \textbf{48.0} & \textbf{21.9} & 6,332 & 17,525 & 326 \\

        \bottomrule
    \end{tabular}}
    \caption{Performance Comparison of Radar and Camera Refined Tracklet Set with Baselines. We incorporate ByteTrack and BOTSORT in our CRAFTBooster), denoted as Ours w/ByteTrack and Ours w/BOTSORT. StrongSORT*: We implement StrongSORT with some modifications: removing AFLink, ECC, and BoT( We replace it with our own detector).}
    \label{tab:CR_Refined}
\end{table}

\begin{figure}[t]
    \centering
    \includegraphics[width=1\linewidth]{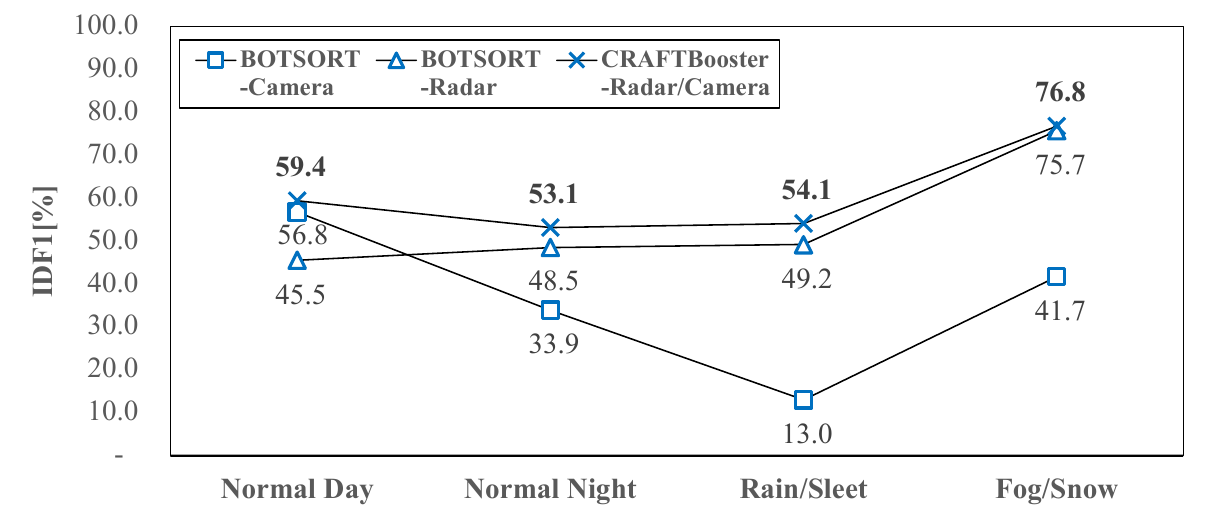}
    \caption{Comparison of IDF1\% under various weather conditions on K-Radar dataset. BOTSORT-Camera and BOTSORT-Radar mean that the tracking result is from BOTSORT with different kind of senso. The online trackers in CRAFTBooster are based on BOTSORT in two modalities.}
    \label{fig:Weather_IDF1}
\end{figure}

In various weather conditions, each sensor modality is affected by lighting and environmental factors, resulting in degradation in performance, as shown in Fig~\ref{fig:Weather_IDF1}. In favorable weather conditions with normal lighting, the camera exhibits superior performance. Conversely, in other adverse environments, radar outperforms the camera. During rain/sleet, the camera often struggles due to blurred images, resulting in the poorest performance. As the result, it is evident that our method enhances performance across various weather conditions, surpassing the best-performing modality in the given environment.

\begin{figure*}[h]
    \centering
    \includegraphics[width=1\linewidth]{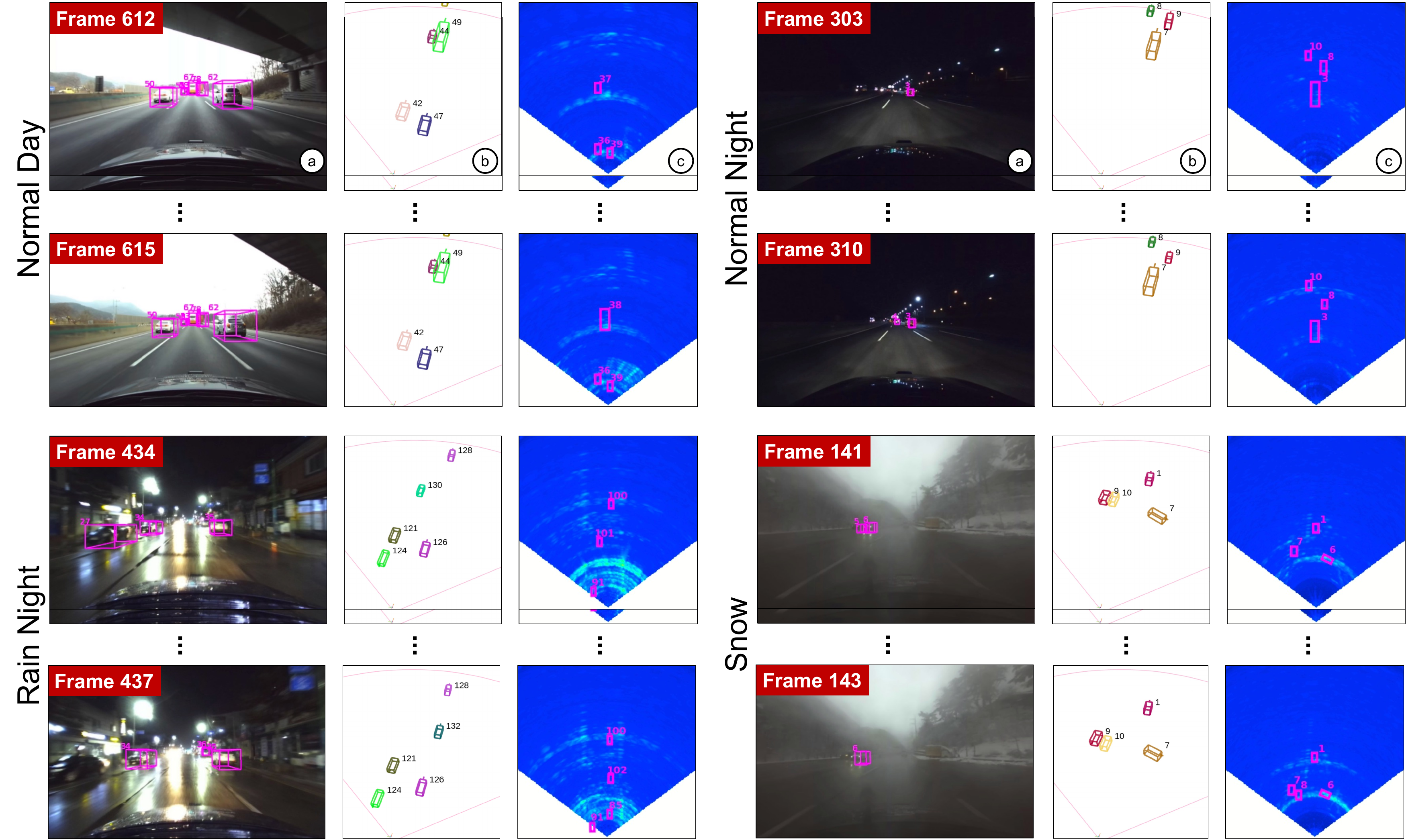}

    \caption{Samples from the result under various weather conditions on K-Radar dataset. We visualize the fusion tracking result(b), comparing with BOTSORT's result on camera(a) and radar(c). }
    \label{fig:vis_exp}
\end{figure*}

Camera detection remains stable under normal conditions in Fig~\ref{fig:vis_exp}, but exhibits issues with miss-detection in inclement weather, confirming in Fig~\ref{fig:Weather_IDF1}. It's worth noting that we found that the primary factors influencing radar results are the density of vehicles or the abundance of surrounding objects, possibly tied to equipment resolution. 


\section{Ablation Study}

In Table~\ref{tab:Ablations_CMCModule}, We analyze the performance enhancement achieved by each individual component within the \textbf{Cross-modality module}. The methods, \textbf{Cross-modality on Unmatched Tracklets} and \textbf{Cross-modality on Unmatched Detections} have been explained in Section 3. The term \textbf{Update} refers to whether the results augmented by these two methods need to be used to update the motion model of each online tracker, influencing the predicted locations in the subsequent frames. The results from using ByteTrack or BOTSORT as online tracker demonstrate that utilizing all methods yields the best outcome.

\begin{table}[t]
    \centering
    \small
    \resizebox{1\linewidth}{!}{
    \begin{tabular}{c|ccc|ccccc}
        \toprule
        Methods&\makecell[c]{Cro.CM\\on Un.Tr}&\makecell[c]{Cro.CM\\on Un.Det} & Update& IDF1$\uparrow$ & MOTA$\uparrow$ & FP$\downarrow$ & FN$\downarrow$ & IDs$\downarrow$ \\
        \midrule
        \multirowcell{5}{Ours \\w/ ByteTrack}&&  && 54.9 & 42.9 & 3,640 & 13,498 & 532\\
        
        &&\checkmark&\checkmark& 54.9 & 43.2 & 3,792 & 13,275  & 537 \\
        &\checkmark&&\checkmark& 57.2 & 43.2 & 3,733 & 13,391 & 468 \\
        &\checkmark&\checkmark&& 56.8 & 43.3 & 3,700 & 13,397 & 469\\
        &\checkmark&\checkmark&\checkmark& \textbf{ 58.3 } & \textbf{ 44.7 } & 3,653 & 13,076 & 400 \\
         \midrule
        \multirowcell{5}{Ours \\w/ BOTSORT}&&  & & 56.2 & 43.6 & 3,624 & 13,369 & 479\\
        
        &&\checkmark&\checkmark& 58.4 & 43.7 & 3,761 & 13,269 & 393\\
        &\checkmark&&\checkmark& 56.5 & 43.9 & 3,731 & 13,195  & 484\\
        &\checkmark&\checkmark&&  58.2  &  44.1 & 3,747 & 13,263 & 385\\
        &\checkmark&\checkmark&\checkmark& \textbf{59.5} & \textbf{44.9} & 3,751 & 12,965 & 356\\

        \bottomrule
    \end{tabular}}
    \caption{Ablations on Cross-modality Check Module. CroCM means Cross-modality Check Module.}
    \label{tab:Ablations_CMCModule}
\end{table}

In Table~\ref{tab:PairingCost}, we analyze which method of pairing cost can get the best results. The methods involving avg. IoU, avg. AoP, and Avg. CD are all evaluated in \textbf{Tracklets Pairing} within \textbf{Cross-modality Check Module}. The results demonstrate that using avg. CD consistently achieves the best performance. We found that there is a significant difference in Avg. AoP, because when tracklets are on the same lane, the AoP values tend to approach zero. This makes it challenging to use as a matching criterion.

\begin{table}[t]
    
    \centering
    \small
    \resizebox{1\linewidth}{!}{
    \begin{tabular}{c|c|ccccc}
        \toprule
        Method&\makecell[c]{Pairing \\Metric}& IDF1$\uparrow$ & MOTA$\uparrow$ & FP$\downarrow$ & FN$\downarrow$ & IDs$\downarrow$ \\
        \midrule
        \multirowcell{3}{Ours\\w/ ByteTrack}&Avg. IoU & 53.2 & 39.4 & 5,011 & 12,954 & 810\\
        &Avg. AoP& 44.0 & 23.8 & 7,780 & 12,813  & 2,993\\
        &Avg. CD& \textbf{ 58.3 } & \textbf{ 44.7 } & 3,653 & 13,076 & 400\\
        \midrule
        \multirowcell{3}{Ours\\w/ BOTSORT}&Avg. IoU & 55.2 & 40.5 & 4,905 & 12,809 & 721\\
        &Avg. AoP& 44.5 & 23.2 & 7,764 & 13,052  & 2,961\\
        &Avg. CD& \textbf{59.5} & \textbf{44.9} & 3,751 & 12,965 & 356\\
        \bottomrule
    \end{tabular}}
    \caption{Ablation Analysis on Pairing Metric}
    \label{tab:PairingCost}
\end{table}

In the \textbf{Tracklets Fusion with Homoscedastic Uncertainty}, we utilize the reciprocal of the standard deviation of location displacement as fusion weights, normalized by the softmax function. Table~\ref{tab:Ablations_Tracklet_Fusion} examines the impact of using uniform weights or softmax. The findings reveal that uniform weights yield inferior performance, whereas employing softmax leads to a slight improvement in performance.

\begin{table}[t]
    \centering
    \small
    \resizebox{1\linewidth}{!}{
    \begin{tabular}{c|c|ccccc}
        \toprule
        Method&\makecell[c]{Fusion \\Method}& IDF1$\uparrow$ & MOTA$\uparrow$ & FP$\downarrow$ & FN$\downarrow$ & IDs$\downarrow$ \\
        \midrule
        \multirowcell{3}{Ours \\w/ ByteTrack}& Uniform Weighted & 57.8 & 44.3 & 3,710 & 13,132 & 411\\
        & Weighted Sum& 58.2 & 44.5 & 3,735 & 13,057  & 395\\
        &Softmax&  \textbf{ 58.3 } & \textbf{ 44.7 } & 3,653 & 13,076 & 400\\
        \midrule
        \multirowcell{3}{Ours \\w/ BOTSORT}& Uniform Weighted & 59.2 & 44.8 & 3,754 & 12,968 & 365\\
        &Weighted Sum& 59.4 & 44.8 & 3,782 & 12,986  & 354\\
        &Softmax& \textbf{59.5} & \textbf{44.9} & 3,751 & 12,965 & 356\\
        \bottomrule
    \end{tabular}}
    \caption{Ablation Analysis on Weights of Paired Tracklet Fusion.}
    \label{tab:Ablations_Tracklet_Fusion}
\end{table}

\section{Limitation}
We believe that the concept presented in this article holds the potential for a long-term research vision. The current method still possess certain limitations and possibilities for further development. The primary challenge confronted by our method is that the fusion result cannot surpass all the refined tracklet set after Cross-modality Check Module, as evidenced by the comparison of results in Table \ref{tab:trackerPerformance} and Table \ref{tab:CR_Refined}. Through the visualization of results, we find that the main issue lies in the quantity of false positives (FPs), which can also be observed in Table \ref{tab:CR_Refined}. The FPs of refined tracklets on each modality are more than the the origional tracker. It remains difficult to determine whether these instances truly exist based on limited information from the other modality. We expect that this issue can be overcome by extracting additional information from the raw data of each modality in the future.

\section{Conclusion}
In the realm of 3D Multiple Object Tracking (MOT), our approach, CRAFTBooster, introduces a novel fusion concept by incorporating the fusion process within the tracking stage. 
We effectively exploit the unique characteristics of both camera and radar sensors by simultaneously evaluating 3D object detections in both camera's perspective view and radar's BEV. 
As a result, perception's unreliability introduced by one modality can be complemented by the other.
Regardless of varying weather conditions, our method compensates for erroneous detections by incorporating complementary information from the other sensor. This optimizes overall performance and ensures system stability.
Furthermore, our method can seamlessly integrate with existing well-performed online trackers.
In the experiments, our approach consistently improves about 5-6\% IDF1 scores on K-Radar and 1-2\% on CRUW3D benchmark. This work not only demonstrates the power of effective sensor fusion but also paves the way for enhanced 3D MOT tasks in real-world scenarios.








\newpage
\bibliography{main}

\begin{thebibliography}{10}\itemsep=-1pt

\bibitem{aharon2022botsort}
Nir Aharon, Roy Orfaig, and Ben-Zion Bobrovsky.
\newblock Bot-sort: Robust associations multi-pedestrian tracking, 2022.

\bibitem{TransFusion}
Xuyang Bai, Zeyu Hu, Xinge Zhu, Qingqiu Huang, Yilun Chen, Hangbo Fu, and Chiew-Lan Tai.
\newblock Transfusion: Robust lidar-camera fusion for 3d object detection with transformers.
\newblock In {\em 2022 IEEE/CVF Conference on Computer Vision and Pattern Recognition (CVPR)}, pages 1080--1089, 2022.

\bibitem{nuscenes}
Holger Caesar, Varun Bankiti, Alex~H. Lang, Sourabh Vora, Venice~Erin Liong, Qiang Xu, Anush Krishnan, Yu Pan, Giancarlo Baldan, and Oscar Beijbom.
\newblock nuscenes: A multimodal dataset for autonomous driving.
\newblock In {\em CVPR}, 2020.

\bibitem{cai2022memot}
Jiarui Cai, Mingze Xu, Wei Li, Yuanjun Xiong, Wei Xia, Zhuowen Tu, and Stefano Soatto.
\newblock Memot: Multi-object tracking with memory.
\newblock In {\em Proceedings of the IEEE/CVF Conference on Computer Vision and Pattern Recognition}, pages 8090--8100, 2022.

\bibitem{cao2023observationcentric}
Jinkun Cao, Jiangmiao Pang, Xinshuo Weng, Rawal Khirodkar, and Kris Kitani.
\newblock Observation-centric sort: Rethinking sort for robust multi-object tracking, 2023.

\bibitem{chai2022deep}
Wenhao Chai and Gaoang Wang.
\newblock Deep vision multimodal learning: Methodology, benchmark, and trend.
\newblock {\em Applied Sciences}, 12(13):6588, 2022.

\bibitem{cheng2023centerradarnet}
Jen-Hao Cheng, Sheng-Yao Kuan, Hugo Latapie, Gaowen Liu, and Jenq-Neng Hwang.
\newblock Centerradarnet: Joint 3d object detection and tracking framework using 4d fmcw radar, 2023.

\bibitem{3d_rob_benchmarking}
Yinpeng Dong, Caixin Kang, Jinlai Zhang, Zijian Zhu, Yikai Wang, Xiao Yang, Hang Su, Xingxing Wei, and Jun Zhu.
\newblock Benchmarking robustness of 3d object detection to common corruptions.
\newblock In {\em Proceedings of the IEEE/CVF Conference on Computer Vision and Pattern Recognition}, pages 1022--1032, 2023.

\bibitem{du2022giaotracker}
Yunhao Du, Junfeng Wan, Yanyun Zhao, Binyu Zhang, Zhihang Tong, and Junhao Dong.
\newblock Giaotracker: A comprehensive framework for mcmot with global information and optimizing strategies in visdrone 2021, 2022.

\bibitem{du2023strongsort}
Yunhao Du, Zhicheng Zhao, Yang Song, Yanyun Zhao, Fei Su, Tao Gong, and Hongying Meng.
\newblock Strongsort: Make deepsort great again.
\newblock {\em IEEE Transactions on Multimedia}, 2023.

\bibitem{ge2021yolox}
Zheng Ge, Songtao Liu, Feng Wang, Zeming Li, and Jian Sun.
\newblock Yolox: Exceeding yolo series in 2021, 2021.

\bibitem{packnet}
Vitor Guizilini, Rares Ambrus, Sudeep Pillai, Allan Raventos, and Adrien Gaidon.
\newblock 3d packing for self-supervised monocular depth estimation.
\newblock In {\em IEEE Conference on Computer Vision and Pattern Recognition (CVPR)}, 2020.

\bibitem{simplebev}
Adam~W. Harley, Zhaoyuan Fang, Jie Li, Rares Ambrus, and Katerina Fragkiadaki.
\newblock Simple-bev: What really matters for multi-sensor bev perception?
\newblock In {\em 2023 IEEE International Conference on Robotics and Automation (ICRA)}, pages 2759--2765, 2023.

\bibitem{yuhong-CMKD-ECCV2022}
Yu Hong, Hang Dai, and Yong Ding.
\newblock Cross-modality knowledge distillation network for monocular 3d object detection.
\newblock In {\em {ECCV}}, Lecture Notes in Computer Science. Springer, 2022.

\bibitem{uniad}
Yihan Hu, Jiazhi Yang, Li Chen, Keyu Li, Chonghao Sima, Xizhou Zhu, Siqi Chai, Senyao Du, Tianwei Lin, Wenhai Wang, et~al.
\newblock Planning-oriented autonomous driving.
\newblock In {\em Proceedings of the IEEE/CVF Conference on Computer Vision and Pattern Recognition}, pages 17853--17862, 2023.

\bibitem{huang2024exploring}
Hsiang-Wei Huang, Cheng-Yen Yang, Wenhao Chai, Zhongyu Jiang, and Jenq-Neng Hwang.
\newblock Exploring learning-based motion models in multi-object tracking.
\newblock {\em arXiv preprint arXiv:2403.10826}, 2024.

\bibitem{Huang2020EPNetEP}
Tengteng Huang, Zhe Liu, Xiwu Chen, and Xiang Bai.
\newblock Epnet: Enhancing point features with image semantics for 3d object detection.
\newblock July 2020.

\bibitem{liang2022bevfusion}
Tingting Liang, Hongwei Xie, Kaicheng Yu, Zhongyu Xia, Zhiwei Lin, Yongtao Wang, Tao Tang, Bing Wang, and Zhi Tang.
\newblock Bevfusion: A simple and robust lidar-camera fusion framework, 2022.

\bibitem{liu2022bevfusion}
Zhijian Liu, Haotian Tang, Alexander Amini, Xinyu Yang, Huizi Mao, Daniela Rus, and Song Han.
\newblock Bevfusion: Multi-task multi-sensor fusion with unified bird's-eye view representation, 2022.

\bibitem{bevfusion}
Zhijian Liu, Haotian Tang, Alexander Amini, Xinyu Yang, Huizi Mao, Daniela~L. Rus, and Song Han.
\newblock Bevfusion: Multi-task multi-sensor fusion with unified bird's-eye view representation.
\newblock In {\em 2023 IEEE International Conference on Robotics and Automation (ICRA)}, pages 2774--2781, 2023.

\bibitem{9607427}
Michael Meyer, Georg Kuschk, and Sven Tomforde.
\newblock Graph convolutional networks for 3d object detection on radar data.
\newblock In {\em 2021 IEEE/CVF International Conference on Computer Vision Workshops (ICCVW)}, pages 3053--3062, 2021.

\bibitem{nabati2021centerfusion}
Ramin Nabati and Hairong Qi.
\newblock Centerfusion: Center-based radar and camera fusion for 3d object detection.
\newblock In {\em Proceedings of the IEEE/CVF Winter Conference on Applications of Computer Vision}, pages 1527--1536, 2021.

\bibitem{paek2023kradar}
Dong-Hee Paek, Seung-Hyun Kong, and Kevin~Tirta Wijaya.
\newblock K-radar: 4d radar object detection for autonomous driving in various weather conditions, 2023.

\bibitem{CNN_RUDet_3DRadarCube}
Andras Palffy, Jiaao Dong, Julian F.~P. Kooij, and Dariu~M. Gavrila.
\newblock Cnn based road user detection using the 3d radar cube.
\newblock {\em IEEE Robotics and Automation Letters}, 5(2):1263--1270, 2020.

\bibitem{CLOCs}
Su Pang, Daniel Morris, and Hayder Radha.
\newblock Clocs: Camera-lidar object candidates fusion for 3d object detection.
\newblock In {\em 2020 IEEE/RSJ International Conference on Intelligent Robots and Systems (IROS)}, pages 10386--10393, 2020.

\bibitem{park2021pseudolidar}
Dennis Park, Rares Ambrus, Vitor Guizilini, Jie Li, and Adrien Gaidon.
\newblock Is pseudo-lidar needed for monocular 3d object detection?, 2021.

\bibitem{schumann2021radarscenes}
Ole Schumann, Markus Hahn, Nicolas Scheiner, Fabio Weishaupt, Julius~F. Tilly, Jürgen Dickmann, and Christian Wöhler.
\newblock Radarscenes: A real-world radar point cloud data set for automotive applications, 2021.

\bibitem{sheeny2021radiate}
Marcel Sheeny, Emanuele~De Pellegrin, Saptarshi Mukherjee, Alireza Ahrabian, Sen Wang, and Andrew Wallace.
\newblock Radiate: A radar dataset for automotive perception in bad weather, 2021.

\bibitem{pvrcnn}
Shaoshuai Shi, Chaoxu Guo, Li Jiang, Zhe Wang, Jianping Shi, Xiaogang Wang, and Hongsheng Li.
\newblock Pv-rcnn: Point-voxel feature set abstraction for 3d object detection.
\newblock In {\em 2020 IEEE/CVF Conference on Computer Vision and Pattern Recognition (CVPR)}, pages 10526--10535, 2020.

\bibitem{song2023moviechat}
Enxin Song, Wenhao Chai, Guanhong Wang, Yucheng Zhang, Haoyang Zhou, Feiyang Wu, Xun Guo, Tian Ye, Yan Lu, Jenq-Neng Hwang, et~al.
\newblock Moviechat: From dense token to sparse memory for long video understanding.
\newblock {\em arXiv preprint arXiv:2307.16449}, 2023.

\bibitem{Vora2019PointPaintingSF}
Sourabh Vora, Alex~H. Lang, Bassam Helou, and Oscar Beijbom.
\newblock Pointpainting: Sequential fusion for 3d object detection.
\newblock {\em 2020 IEEE/CVF Conference on Computer Vision and Pattern Recognition (CVPR)}, pages 4603--4611, 2019.

\bibitem{cruw3d}
Yizhou Wang, Jen-Hao Cheng, Jui-Te Huang, Sheng-Yao Kuan, Qiqian Fu, Chiming Ni, Shengyu Hao, Gaoang Wang, Guanbin Xing, Hui Liu, and Jenq-Neng Hwang.
\newblock Vision meets mmwave radar: 3d object perception benchmark for autonomous driving.
\newblock {\em arXiv preprint}, 2023.

\bibitem{rodnet}
Yizhou Wang, Zhongyu Jiang, Yudong Li, Jenq-Neng Hwang, Guanbin Xing, and Hui Liu.
\newblock Rodnet: A real-time radar object detection network cross-supervised by camera-radar fused object 3d localization.
\newblock {\em IEEE Journal of Selected Topics in Signal Processing}, 15(4):954--967, 2021.

\bibitem{centerpoint}
Tianwei Yin, Xingyi Zhou, and Philipp Krähenbühl.
\newblock Center-based 3d object detection and tracking.
\newblock In {\em 2021 IEEE/CVF Conference on Computer Vision and Pattern Recognition (CVPR)}, pages 11779--11788, 2021.

\bibitem{8578353}
Fisher Yu, Dequan Wang, Evan Shelhamer, and Trevor Darrell.
\newblock Deep layer aggregation.
\newblock In {\em 2018 IEEE/CVF Conference on Computer Vision and Pattern Recognition}, pages 2403--2412, 2018.

\bibitem{fusion_rob_benchmarking}
Kaicheng Yu, Tang Tao, Hongwei Xie, Zhiwei Lin, Tingting Liang, Bing Wang, Peng Chen, Dayang Hao, Yongtao Wang, and Xiaodan Liang.
\newblock Benchmarking the robustness of lidar-camera fusion for 3d object detection.
\newblock In {\em Proceedings of the IEEE/CVF Conference on Computer Vision and Pattern Recognition}, pages 3187--3197, 2023.

\bibitem{zhang2022bytetrack}
Yifu Zhang, Peize Sun, Yi Jiang, Dongdong Yu, Fucheng Weng, Zehuan Yuan, Ping Luo, Wenyu Liu, and Xinggang Wang.
\newblock Bytetrack: Multi-object tracking by associating every detection box, 2022.

\bibitem{Zheng_2022}
Lianqing Zheng, Zhixiong Ma, Xichan Zhu, Bin Tan, Sen Li, Kai Long, Weiqi Sun, Sihan Chen, Lu Zhang, Mengyue Wan, Libo Huang, and Jie Bai.
\newblock {TJ}4dradset: A 4d radar dataset for autonomous driving.
\newblock In {\em 2022 {IEEE} 25th International Conference on Intelligent Transportation Systems ({ITSC})}. {IEEE}, oct 2022.

\bibitem{zheng2019distanceiou}
Zhaohui Zheng, Ping Wang, Wei Liu, Jinze Li, Rongguang Ye, and Dongwei Ren.
\newblock Distance-iou loss: Faster and better learning for bounding box regression, 2019.

\bibitem{Zhu_2023_CVPR}
Zijian Zhu, Yichi Zhang, Hai Chen, Yinpeng Dong, Shu Zhao, Wenbo Ding, Jiachen Zhong, and Shibao Zheng.
\newblock Understanding the robustness of 3d object detection with bird's-eye-view representations in autonomous driving.
\newblock In {\em Proceedings of the IEEE/CVF Conference on Computer Vision and Pattern Recognition (CVPR)}, pages 21600--21610, June 2023.

\end{thebibliography}

\end{document}